%% file: main.tex
\documentclass{article} 
\usepackage{iclr2026_conference,times}

\input{math_commands.tex}

\usepackage{hyperref}
\usepackage{url}
\usepackage{graphicx}   
\usepackage{booktabs,multirow,makecell} 
\usepackage[table]{xcolor}
\usepackage{subcaption}
\usepackage{wrapfig} 
\usepackage{enumitem}
\setlist[itemize]{topsep=3pt, partopsep=0pt, parsep=0pt, itemsep=3pt}

\title{PEAR: Phase Entropy Aware Reward for Efficient Reasoning}

\author{
Chen Huang\textsuperscript{1} \quad 
Wei Lu\textsuperscript{2} \quad 
Wenxuan Zhang\textsuperscript{1} \\ 
\textsuperscript{1}Singapore University of Technology and Design \quad
\textsuperscript{2}Nanyang Technological University \\
\texttt{chen\_huang@mymail.sutd.edu.sg} \quad
\texttt{wei.lu@ntu.edu.sg} \\
\texttt{wxzhang@sutd.edu.sg} 
}

\iclrfinalcopy 
\begin{document}

\maketitle

\begin{abstract}
Large Reasoning Models (LRMs) have achieved impressive performance on complex reasoning tasks by generating detailed chain-of-thought (CoT) explanations. 
However, these responses are often excessively long, containing redundant reasoning steps that inflate inference cost and reduce usability. 
Controlling the length of generated reasoning without sacrificing accuracy remains an open challenge. 
Through a systematic empirical analysis, we reveal a consistent positive correlation between model entropy and response length at different reasoning stages across diverse LRMs: 
the thinking phase exhibits higher entropy, reflecting exploratory behavior of longer responses, while the final answer phase shows lower entropy, indicating a more deterministic solution.
This observation suggests that entropy at different reasoning stages can serve as a control knob for balancing conciseness and performance. 
Based on this insight, this paper introduces \textbf{P}hase \textbf{E}ntropy \textbf{A}ware \textbf{R}eward (PEAR), a reward mechanism that incorporating phase-dependent entropy into the reward design.  
Instead of treating all tokens uniformly, PEAR penalize excessive entropy during the thinking phase and allowing moderate exploration at the final answer phase, which encourages models to generate concise reasoning traces that retain sufficient flexibility to solve the task correctly. 
This enables adaptive control of response length without relying on explicit length targets or rigid truncation rules. 
Extensive experiments across four benchmarks demonstrate that PEAR consistently reduces response length while sustaining competitive accuracy across model scales. 
In addition, PEAR demonstrates strong out-of-distribution (OOD) robustness beyond the training distribution. Our code is available at: \url{https://github.com/iNLP-Lab/PEAR}.
\end{abstract}

\input{Sec/01_Introduction}

\input{Sec/03_Preliminary_Analysis}
\input{Sec/04_Method}
\input{Sec/06_Results}

\input{Sec/02_Related_Work_Concise}
\input{Sec/07_Conclusion}
\clearpage

\bibliography{custom}
\bibliographystyle{iclr2026_conference}
\clearpage

\appendix
\input{Appendix/02_Qwen3_8B_Entropy_filter_results}

\input{Appendix/03_Experiment_Details_for_baselines}

\input{Appendix/04_Evaluation_Benchmarks}

\end{document}

%% file: math_commands.tex
\usepackage{amsmath,amsfonts,bm}

\def\eqref#1{equation~\ref{#1}}

\def\1{\bm{1}}

\DeclareMathAlphabet{\mathsfit}{\encodingdefault}{\sfdefault}{m}{sl}
\SetMathAlphabet{\mathsfit}{bold}{\encodingdefault}{\sfdefault}{bx}{n}



%% file: Sec/01_Introduction.tex
\section{Introduction}

Large Language Models (LLMs) have demonstrated remarkable reasoning capabilities, particularly when employing techniques like Chain-of-Thought (COT) prompting \citep{wei2022chain}. Building on this, recent Large Reasoning Models (LRMs) \citep{jaech2024openai, guo2025deepseek, yang2025qwen3, team2025kimi, qwq32b} encourage an explicit thinking phase via special tokens before generating the final answer, further improving models' complex problem-solving capability.
However, LRMs tend to generate excessively long chain-of-thought responses~\citep{chen2024not, yue2025don}, the models often produce redundant calculations or verbose explanations, which leads to bloated outputs and reduces inference efficiency \citep{hassid2025don, kuo2025h}. 
Consequently, a key challenge is to enable models to think less while preserving the performance.

\begin{figure}[t] 
  \centering
  \includegraphics[width=0.83\columnwidth]{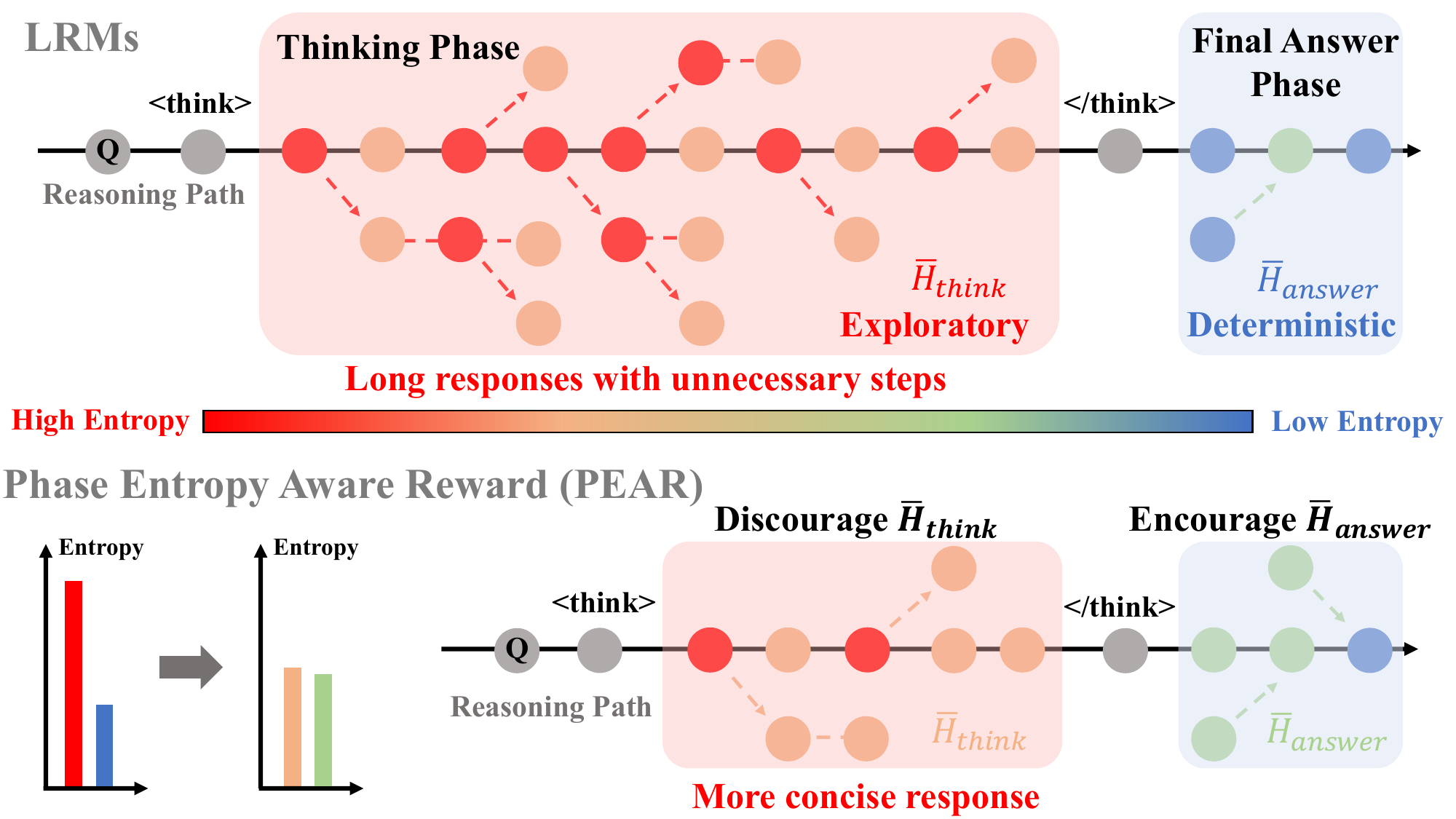} 
  \caption{PEAR reduce the response length by penalizing excessive entropy during thinking phase while allowing moderate exploration at the final answer phase.}
  \label{fig:method}
  \vspace{-10mm}
\end{figure}

Recent works have attempted to address this issue by enforcing efficiency through further training on filtered concise data \citep{yue2025don, qu2025survey,sui2025stop}. 
The common paradigm is to modify the training corpus so that the model is exposed primarily to shorter reasoning traces \citep{yuan2025not, an2025don, zhao2025can}. 
By strictly constraining the supervision signal, the model often struggles to adapt to novel reasoning styles or out-of-distribution (OOD) problems where the optimal length of reasoning may differ \citep{aggarwal2025l1}. 
Moreover, such methods risk discarding valuable intermediate reasoning that could improve accuracy. 
This motivates the need for a more adaptive and model-driven approach to efficient reasoning.

Concurrently, there has been growing interest in understanding how token-level uncertainty, as measured by entropy, influences model behavior \citep{lei2025revisiting, cheng2025reasoning, zhang-etal-2025-entropy}. 
Entropy captures the spread of the predictive distribution: 
high-entropy segments often correspond to exploratory reasoning steps where the model searches for a correct path, while low-entropy segments capture more deterministic computations or final answer generation \citep{wang2025beyond, zhang2025no}. 
Therefore, recent works have begun to exploit these signals for improving calibration or enhancing reasoning robustness \citep{zhang2025right, wang2025stabilizing}. 
However, the connection between entropy and efficient reasoning has been largely overlooked. 

Intuitively, a model that operates at consistently high entropy may explore too broadly and thus produce unnecessarily long reasoning chains, while a model biased toward low entropy may commit earlier to a determined reasoning path with more concise outputs. Motivated by this hypothesis, we first conduct empirical analysis, and observe a consistent positive correlation between average token-level entropy and response length across model scales and benchmarks. Interestingly, this relationship is not uniform across reasoning stages: the ``thinking'' portion of the output exhibits substantially higher entropy than the ``final answer'' portion, highlighting distinct roles of exploration and commitment in different stages of reasoning. Moreover, when we filter out high-entropy tokens, models' performance will not be affected within certain ratio, suggesting that excessive entropy can be pruned without harming reasoning quality.
Based on these observations, we propose \textbf{P}hase \textbf{E}ntropy \textbf{A}ware \textbf{R}eward (\textbf{PEAR}), a reward mechanism that explicitly decomposes entropy into thinking and final answer phases and integrates both components into the training objective.
As illustrated in Figure~\ref{fig:method}, by penalizing excessive entropy during the thinking phase while moderating entropy in the final answer phase, PEAR encourages models to produce more concise reasoning traces, providing a soft and adaptive mechanism for balancing exploration with efficiency.

We evaluate PEAR on four widely used reasoning benchmarks: GSM8K, MATH500, AIME24, and AMC23. 
Across models of different scales, PEAR achieves substantial reductions in response length, ranging from 37.8\% to 59.4\%, while preserving accuracy with decreases of less than 1\%.
By incorporating both phases of a model’s response into the reward calculation, PEAR eliminates the need for manual data curation and generalizes effectively to out-of-domain questions through its broadly applicable training objective.

To summarize, our work makes the following key contributions:
\begin{itemize}
    \item We empirically establish and validate a positive correlation between model entropy and response length in LRMs, and show that the thinking phase exhibits substantially higher entropy than the final answer phase.
    \item We introduce Phase Entropy Aware Reward (PEAR), a reward mechanism that leverages this property to adaptively promote concise reasoning traces without depending on curated datasets or explicit length constraints.
    \item We provide extensive experimental evidence on GSM8K, MATH500, AIME24, and AMC23, showing that our method achieves substantial reductions in response length while preserving accuracy, with strong generalization capability to out-of-distribution tasks.
\end{itemize}

%% file: Sec/03_Preliminary_Analysis.tex
\section{Preliminary Analysis}
\label{sec:preliminary_analysis}

In this section, we present empirical observations that motivate our approach. 
We first examine the relationship between entropy and response length, showing how higher entropy is associated with longer reasoning traces. Next, we differentiate the roles of entropy in the thinking phase versus the final answer phase, highlighting distinct patterns across stages. Finally, we conduct entropy-filtering experiments to demonstrate the robustness of low-entropy reasoning traces. 
All analyses are performed on GSM8K, MATH500, AIME24, and AMC23, where we report average accuracy, response length (in tokens), and entropy.

\subsection{Entropy and Response Length}

\begin{figure}[t]
  \centering
  \begin{minipage}{0.62\linewidth}
    \centering
    \includegraphics[width=\linewidth]{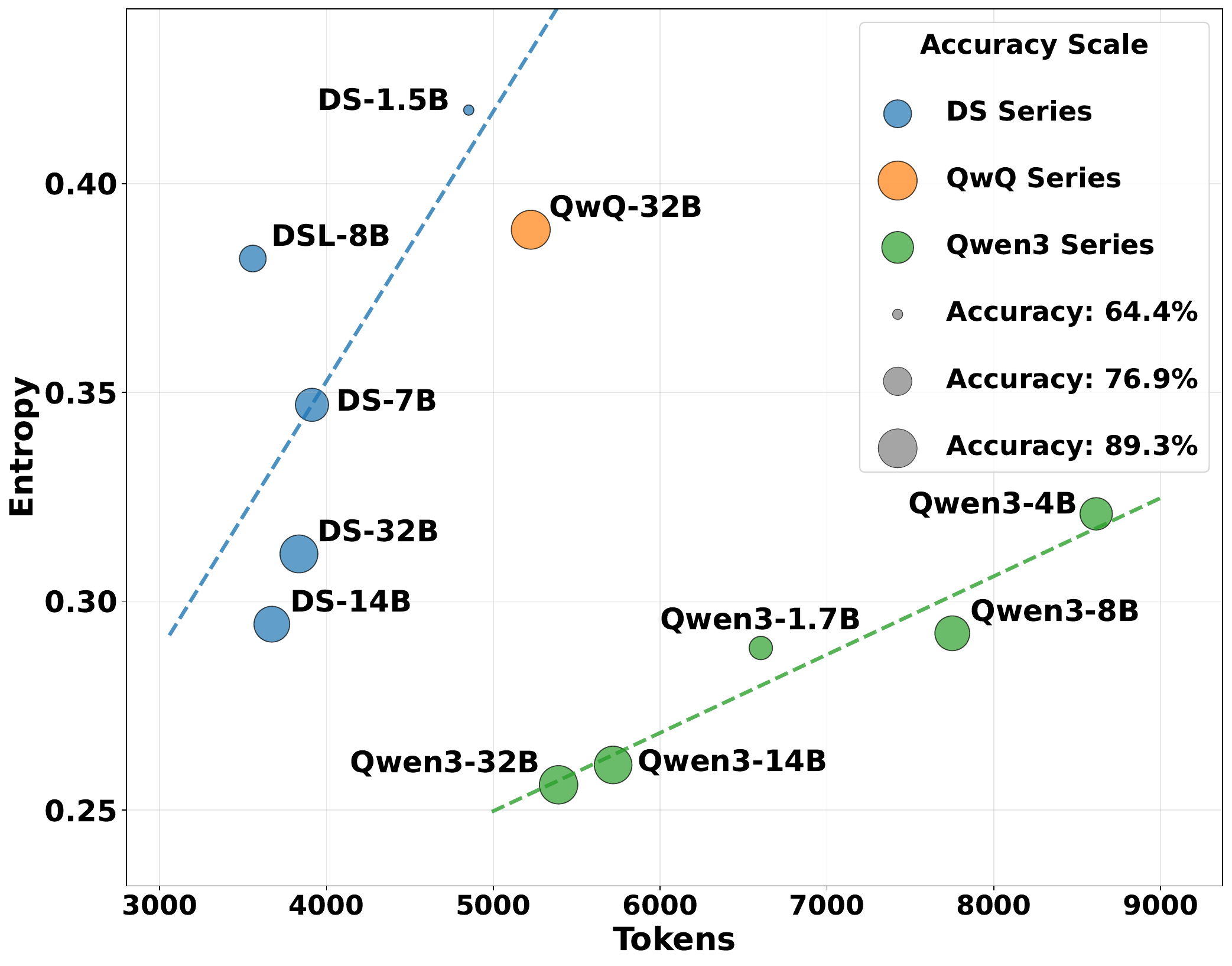}
  \end{minipage}
  \hfill
  \begin{minipage}{0.36\linewidth}
    \centering
    \includegraphics[width=\linewidth]{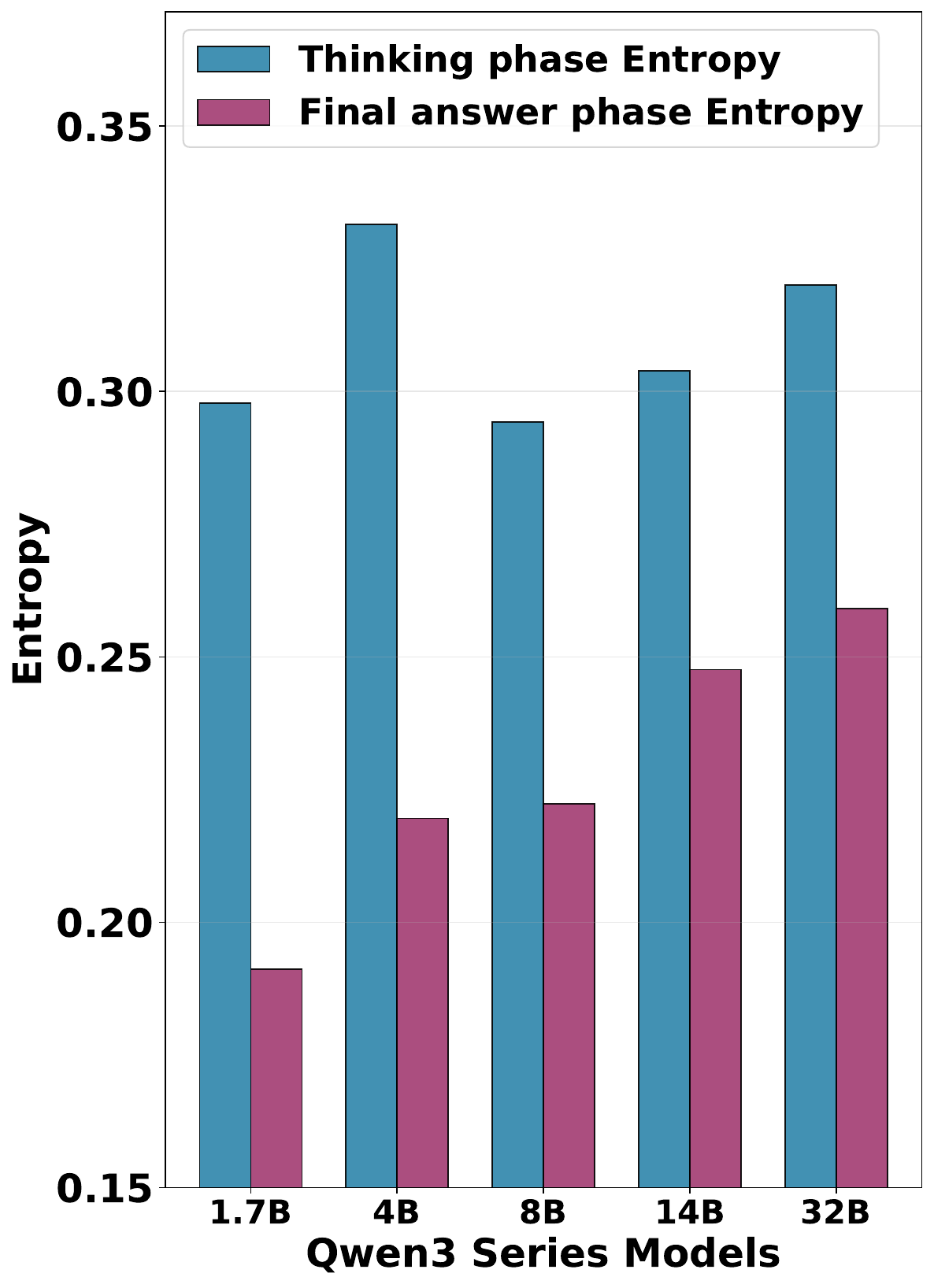}
  \end{minipage}
  \caption{(a) Relationship between average entropy and response length across different models. The dot size indicates accuracy. DS(L) represents DeepSeek-R1-Distill-Qwen/Llama. (b) Comparison of average entropy between the thinking phase and the final answer phase.}
  \label{fig:entropy_inspection}
  \vspace{-5mm}
\end{figure}
We begin by analyzing the correlation between response entropy and length across a diverse set of LRMs. For each model, we measure the average entropy of the predictive distribution across all generated tokens and compare it against the total number of tokens produced during inference.

The entropy of the predictive distribution at each token position $t$ is defined as
\begin{align}
H_t &= -\sum_{i=1}^{|V|} p_i^{(t)} \log p_i^{(t)},
&
\bar{H} &= \frac{1}{T} \sum_{t=1}^{T} H_t
\end{align}
where $p_i^{(t)}$ denotes the predicted probability of token $i$ at position $t$, $|V|$ is the vocabulary size, 
$T$ is the total response length, and $\bar{H}$ is the average entropy across the entire response.

Figure~\ref{fig:entropy_inspection}(a) shows a consistent positive correlation between average entropy and response length across all examined model families and benchmarks. 
Responses with higher entropy are typically longer and more exploratory, while lower entropy corresponds to shorter and more concise traces. 
This pattern is especially evident within individual model series, where models of different scales exhibit a clear alignment between entropy levels and response characteristics. 

These findings suggest that the entropy–length relationship is a fundamental property of large reasoning models. 
Longer responses naturally reflect higher uncertainty or diversity in token predictions, as captured by increased entropy. 
This makes entropy an interpretable internal signal for shaping model behavior. By integrating entropy into the reward design, we can provide models with a principled mechanism to balance thorough reasoning with efficient generation, enabling finer control over response length without relying on explicit constraints.

\subsection{Phase-Dependent Entropy Analysis}
\label{sec:phase-dependent entropy analysis}
To further investigate the role of entropy in model responses, we analyze how entropy is distributed across different stages of generation. 
As shown in Figure~\ref{fig:entropy_inspection}(b), a clear distinction emerges between the thinking phase (before the \texttt{</think>} token) and the final answer phase (after the \texttt{</think>} token). 
The thinking phase exhibits consistently higher entropy, reflecting exploratory behavior as the model searches through multiple potential reasoning paths and generates longer, more diverse traces. 
In contrast, the final answer phase shows much lower entropy, indicating a more confident and deterministic commitment to a specific solution. 
These results indicate that the two phases serve complementary functions of exploration versus conclusion and should therefore be optimized differently. 
Phase-specific reward mechanisms can leverage this distinction, reducing unnecessary exploration during reasoning while preserving confidence and clarity in final answers.

\subsection{Entropy Filtering Experiments}
\label{sec:entropy_filter}

\begin{wrapfigure}{r}{0.5\linewidth} 
  \vspace{-5mm} 
  \centering
  \includegraphics[width=0.95\linewidth]{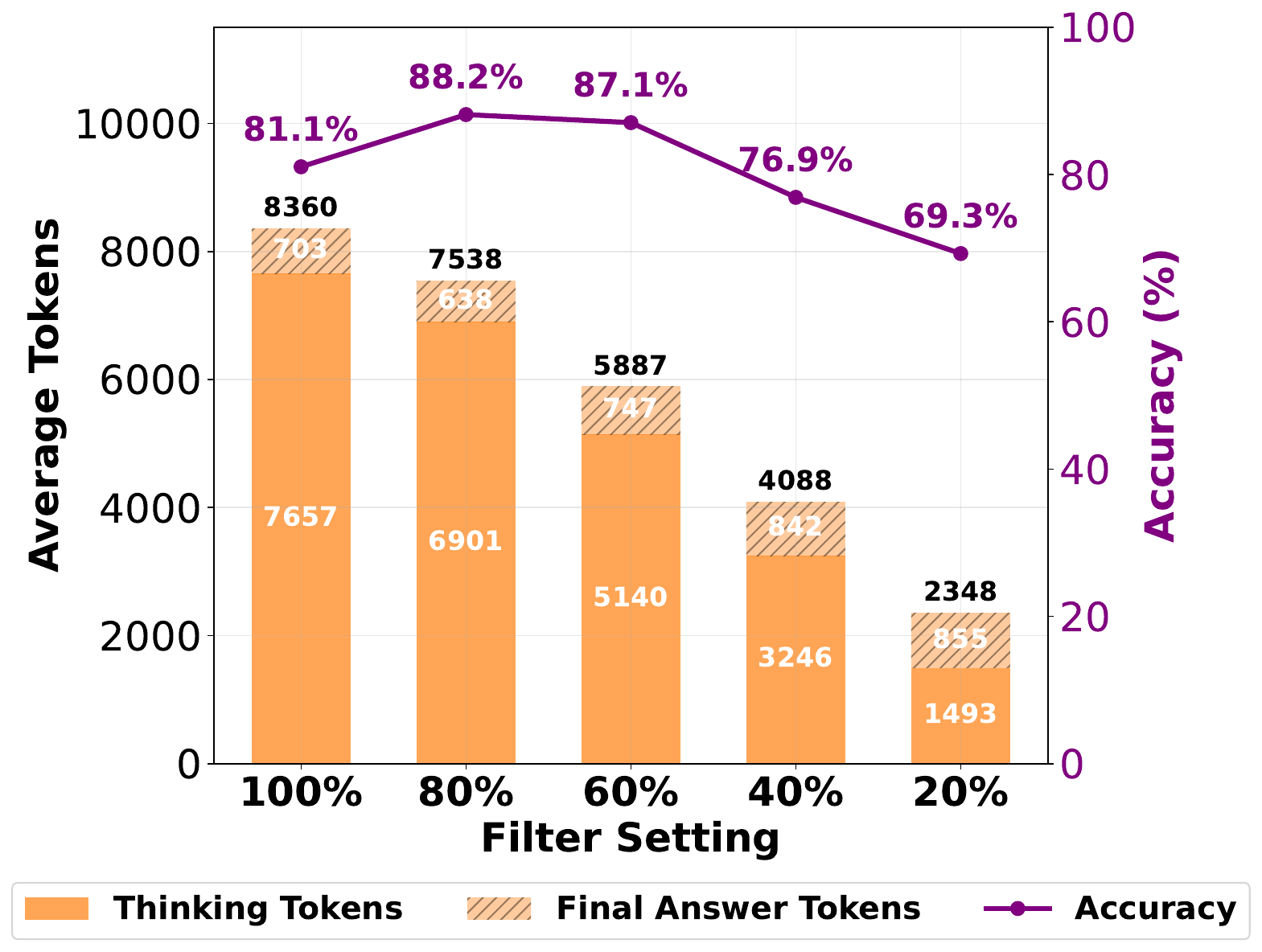}
  \caption{Accuracy and average response length in the entropy filtering experiments on Qwen3-4B.}
  \label{fig:entropy_token_filter_Qwen3_4B}
  \vspace{-5mm} 
\end{wrapfigure}
 
To assess how high-entropy tokens influence model reasoning and whether pruning them impacts reasoning quality, we conduct a systematic filtering experiment, as shown in Figure~\ref{fig:entropy_token_filter_Qwen3_4B}.
Our procedure consists of two stages: first, we generate complete reasoning traces and compute token-level entropy within the thinking phase. 
Second, we retain only a specified percentage of tokens with the lowest entropy values while discarding the rest, thereby constructing filtered reasoning traces. 
These filtered traces are then fed back to the model to produce final answers, enabling us to directly examine how entropy-based filtering influences both reasoning efficiency and task accuracy.
Results for more models can be found at Appendix~\ref{app:entrop_filter_Qwen3_8B}.

When retaining 80\% or 60\% of low-entropy tokens, accuracy remains stable or even improves compared to the unfiltered baseline. 
This indicates that the high-entropy tokens being removed mainly drive excessive exploration rather than contributing to correct reasoning, and their absence reduces noise in the reasoning process. 
Performance degradation only emerges under more aggressive filtering: retaining 40\% or fewer low-entropy tokens leads to a sharp drop in accuracy, showing that essential reasoning steps are lost when the trace is compressed too heavily. 
Notably, the length of the final answer phase remains relatively unchanged across filtering levels, reinforcing that redundancy is concentrated in the thinking phase, where high-entropy tokens leads to over-elaboration and inflates response length without improving outcomes.

%% file: Sec/04_Method.tex
\section{Method}
\subsection{Group Relative Policy Optimization (GRPO)}

We begin with a brief introduction to the Group Relative Policy Optimization (GRPO) algorithm \citep{shao2024deepseekmath}. 
Unlike standard PPO \citep{schulman2017proximal}, 
GRPO eliminates the need for a critic model by estimating advantages through reward normalization across a group of sampled responses to the same prompt.
Specifically, for a prompt $q$ with $G$ responses and corresponding rewards $\{r_i\}_{i=1}^G$, the group-normalized advantage is defined as:
\begin{equation}
\hat{A}_{i,t} = \frac{r_i - \mathrm{mean}(\{r_j\}_{j=1}^G)}{\mathrm{std}(\{r_j\}_{j=1}^G)}.
\label{eq:grpo_advantage}
\end{equation}

This normalization emphasizes the differences among candidate outputs for the same question, which improves the stability of the gradient signal even under sparse reward settings. 
GRPO also incorporates a KL divergence term that regularizes the learned policy against a reference policy. 
The overall surrogate objective can be written as:
\begin{equation}
\begin{aligned}
\mathcal{J}_{\text{GRPO}}(\theta) = 
&\;\mathbb{E}_{\substack{q \sim P(Q),  \{o_i\}_{i=1}^G \sim \pi_{\theta_{\text{old}}}(\cdot|q)}} \\
\frac{1}{G} \sum_{i=1}^G \frac{1}{|o_i|} \sum_{t=1}^{|o_i|}
&\;\; \Bigg\{ \min \Big[ r_{i,t}(\theta) \hat{A}_{i,t},\;
\text{clip }\!\big(r_{i,t}(\theta), 1-\epsilon, 1+\epsilon\big) \hat{A}_{i,t} \Big] 
- \beta D_{\text{KL}}\!\big[\pi_\theta \,\|\, \pi_{\text{ref}}\big] \Bigg\}.
\end{aligned}
\label{eq:grpo_objective}
\end{equation}
where 
\begin{equation}
r_{i,t}(\theta) = \frac{\pi_\theta(o_{i,t} \mid q, o_{i,<t})}{\pi_{\theta_{\text{old}}}(o_{i,t} \mid q, o_{i,<t})},
\end{equation}
$\epsilon$ and $\beta$ are hyperparameters, and $D_{\text{KL}}$ denotes the KL divergence between the learned policy $\pi_\theta$ and a reference policy $\pi_{\text{ref}}$.

\subsection{Phase Entropy Aware Reward (PEAR)}
\label{sec:entropy_reward}

In the original GRPO algorithm, the reward $r$ is typically defined in a rule-based manner, assigning a value of 1 to correct responses and 0 to incorrect ones. 
While simple and effective, this binary scheme overlooks richer characteristics of the response, such as the degree of exploration or reflection embedded in the reasoning trajectory. 
As a result, it provides no guidance on how the model should balance exploratory reasoning with concise and reliable answer generation.

Building on the observed correlation between model entropy and response length in Section~\ref{sec:preliminary_analysis}, we introduce \textbf{Phase Entropy Aware Reward (PEAR)} that leverages entropy as guidance to train models to reason more efficiently.
Let a sampled response be the token sequence
$y=(y_1,\dots,y_T)$ that contains a thinking segment
between \texttt{<think>} and \texttt{</think>} followed by the final answer.

Let $k$ denote the index of the closing token \texttt{</think>} in $y$. 
We compute token entropies with respect to the old policy $\pi_{\theta_{\text{old}}}$:
\begin{equation}
H_t \;=\; - \sum_{v \in \mathcal{V}} \pi_{\theta_{\text{old}}}(v \mid y_{<t})
\log \pi_{\theta_{\text{old}}}(v \mid y_{<t}),
\qquad t=1,\dots,T.
\label{eq:token_entropy}
\end{equation}
We then average entropies for the thinking phase and final answer phase (excluding the \texttt{</think>} token itself):
\begin{align}
\bar{H}_{\text{think}}  &= \frac{1}{k-1}\sum_{t=1}^{k-1} H_t,
&
\bar{H}_{\text{answer}} &= \frac{1}{T-k}\sum_{t=k+1}^{T} H_t.
\end{align}

The phase reward $\mathcal{P}$ integrates entropy from both the thinking and final answer phases, defined as:
\begin{equation}
\mathcal{P}(y) \;=\; \max\!\bigl(0,\; \bar{H}_{\text{think}} - \alpha\,\bar{H}_{\text{answer}}\bigr).
\label{eq:penalty}
\end{equation}
The coefficient \(\alpha\) is a tunable hyperparameter that adjusts the contribution of the final answer phase entropy, enabling flexible control over the balance between reasoning exploration and final answer confidence.
As discussed in Section~\ref{sec:phase-dependent entropy analysis}, the reasoning process exhibits distinct entropy patterns: the thinking phase is characterized by higher entropy with exploratory behavior, while the final answer phase reflects lower entropy associated with deterministic solutions. To promote more efficient reasoning, we therefore aim to reduce entropy during the thinking phase to mitigate unnecessary exploration while preserving or even encouraging entropy in the final answer phase to maintain flexibility and completeness in solution formulation.

Given a base score $s\!\in\!(0,1]$ for a correct final answer and a format score
$r_{\mathrm{fmt}}\!\in\![0,1)$ for malformed/incorrect answers, the phase-aware entropy-inclusive reward for response $y$ is:
\begin{equation}
r(y) \;=\;
\begin{cases}
\min\!\bigl(1,\; s - \mathcal{P}(y)\bigr), & \text{if the extracted answer equals the ground truth},\\[2pt]
r_{\mathrm{fmt}}, & \text{otherwise.}
\end{cases}
\label{eq:entropy_reward}
\end{equation}

Finally, we replace $r_i$ in Eq.~\eqref{eq:grpo_advantage} by $r(y_i)$ and keep
the same GRPO advantage normalization:
\begin{equation}
A_i \;=\; \frac{r(y_i) - \mathrm{mean}\bigl(\{r(y_j)\}_{j=1}^{G}\bigr)}
                 {\mathrm{std}\bigl(\{r(y_j)\}_{j=1}^{G}\bigr)}.
\label{eq:entropy_adv}
\end{equation}

\paragraph{Edge cases.}
If \texttt{</think>} token is absent we set $k\!=\!T$ and use
$\bar{H}_{\text{post}}=0$ (i.e., only thinking phase entropy contributes); if the answer cannot be parsed, we assign $r(y)=r_{\mathrm{fmt}}$.

With PEAR, the model is guided not only by final answer correctness but also by the quality of its reasoning behavior. 
The component for the thinking phase discourages excessive exploration, as high-entropy reasoning yields lower reward, thereby encouraging the model to generate more focused and efficient reasoning traces. 
Meanwhile, the component for the final answer phase helps stabilize and structure the concluding steps, ensuring that the model produces complete and coherent answers without sacrificing accuracy.

%% file: Sec/06_Results.tex
\section{Results}
\label{sec:results}

\subsection{Experiment Setting}
\label{sec:experiment_setting}
\textbf{Baseline Methods.}
\textbf{GRPO} (Group Relative Policy Optimization) \citep{shao2024deepseekmath} is a reinforcement learning framework that eliminates the need for a critic model by estimating advantages through reward normalization within a group of responses to the same prompt. 
\textbf{Step Entropy} \citep{li2025compressing} adopts a two-stage training strategy that enables LLMs to generate compressed chain-of-thought (CoT) reasoning at inference time by strategically inserting [SKIP] tokens. 
\textbf{LCPO} (Length-Controlled Policy Optimization) \citep{aggarwal2025l1} is a reinforcement learning method designed to jointly optimize for accuracy and compliance with user-specified length constraints.

\textbf{Baseline Models.} 
We evaluate our method on widely used Large Reasoning Models (LRMs), including DeepSeek-R1-Distill-Qwen-1.5B \citep{guo2025deepseek}, Qwen3-4B, and Qwen3-8B \citep{yang2025qwen3}, which are commonly adopted in prior works.
For fair comparison, we also report results on these baseline models across different model scales. Detailed implementation settings for all baseline methods are provided in Appendix~\ref{app:baseline method}.

\textbf{Training and Evaluation Setup.} 
We conduct training using the open-source \texttt{verl} framework \citep{sheng2025hybridflow}, 
with 7,473 samples from GSM8K \citep{cobbe2021training} as the training dataset for all models. 
The dataset is consist of grade school math word problems, which are designed to evaluate question answering on basic mathematics that requires multi-step reasoning.
The training configuration uses a batch size of 128 and a learning rate of $1\times 10^{-6}$. We set the coefficient $\alpha$ for final answer phase reward calculation as 1.
To evaluate the effectiveness and generalizability of our compression method, we benchmark on four standard mathematical reasoning datasets: \textbf{GSM8K test set} \citep{cobbe2021training}, \textbf{MATH500} \citep{hendrycks2021measuring}, \textbf{AIME24} \citep{li2024numinamath} and \textbf{AMC23} \citep{li2024numinamath}, detailed introduction of these benchmarks can be found at Appendix~\ref{app:benchmarks}.

Performance is measured along two dimensions: Accuracy (Acc) and the number of Generated Tokens (Tok), with a generation length cap of 16{,}384 tokens. 
Following the evaluation protocol of \citet{guo2025deepseek}, we adopt sampling with temperature set to 0.6 and top-p set to 0.95. 
Answer extraction and verification are carried out following the methodology of \citet{yang2024qwen2}.

\subsection{Effectiveness of PEAR}
\input{Tables/main_results_clear}

As shown in Table~\ref{tab:main_results_clear}, PEAR achieves the most substantial reduction in response length across all benchmarks and evaluated models, while maintaining accuracy at a level comparable to original models. 
Compared to original reasoning models, PEAR achieves an average response length reduction of 37.8\% to 55.2\%, while preserving the same performance with the decrease of only 0.9\% in accuracy.
This indicates that encouraging models to lower entropy level at thinking phase during training provides an effective mechanism for eliminating redundant reasoning steps, thereby producing more concise outputs without compromising correctness. 

Compared to the 1.5B model, the results for the 4B and 8B models suggest that larger models, which are prone to verbose reasoning, benefit more from PEAR by achieving over 50\% reduction in response length. This supports the intuition that bigger models tend to ``over-explain'', creating greater opportunities for efficiency gains. 
Moreover, PEAR delivers a superior efficiency-accuracy trade-off on larger models relative to other baselines. In the case of Qwen3-8B, while Step Entropy and LCPO enforce shorter responses, they incur larger accuracy drops of 1.23\% and 2.68\%, respectively. In contrast, PEAR achieves even greater compression while limiting performance decline to just 0.91\%. This underscores PEAR's adaptive nature, enabling it to compress reasoning traces aggressively without compromising accuracy.

In addition, the benefits of PEAR extend beyond the training distribution, demonstrating strong out-of-distribution (OOD) robustness. Although trained solely on GSM8k, our method yields consistent improvements across all four benchmarks. For example, on Qwen3-4B, PEAR matches the vanilla model’s accuracy on AIME24 and AMC23 while consuming only 34\% and 45\% of the original reasoning budget, respectively. These results highlight that phase-dependent entropy serves as a universal, domain-agnostic signal for controlling reasoning efficiency, enabling our approach to generalize effectively across diverse reasoning tasks.

Overall, these results validate the central hypothesis of our work: incorporating phase-dependent entropy into the reward design enables LRMs to generate shorter and more efficient reasoning trajectories, while preserving accuracy and demonstrating strong generalization across domains.

\subsection{How PEAR affects reasoning}

\begin{figure}[t]
  \centering
  \begin{minipage}{0.495\linewidth}
    \centering
    \includegraphics[height=0.25\textheight]{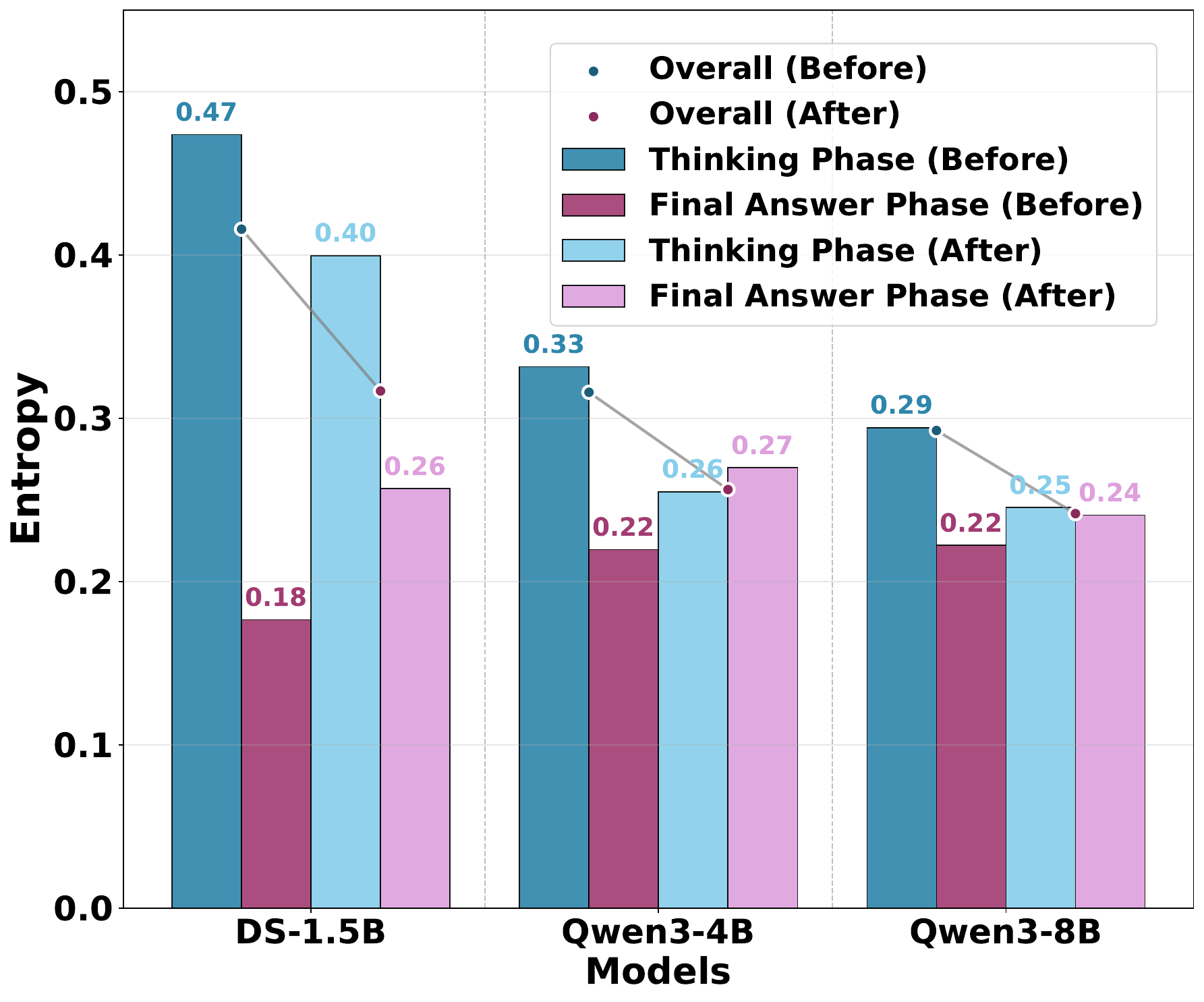}
    \label{fig:entropy_token_changes}
  \end{minipage}%
  \hfill
  \begin{minipage}{0.485\linewidth}
    \centering
    \includegraphics[height=0.245\textheight]{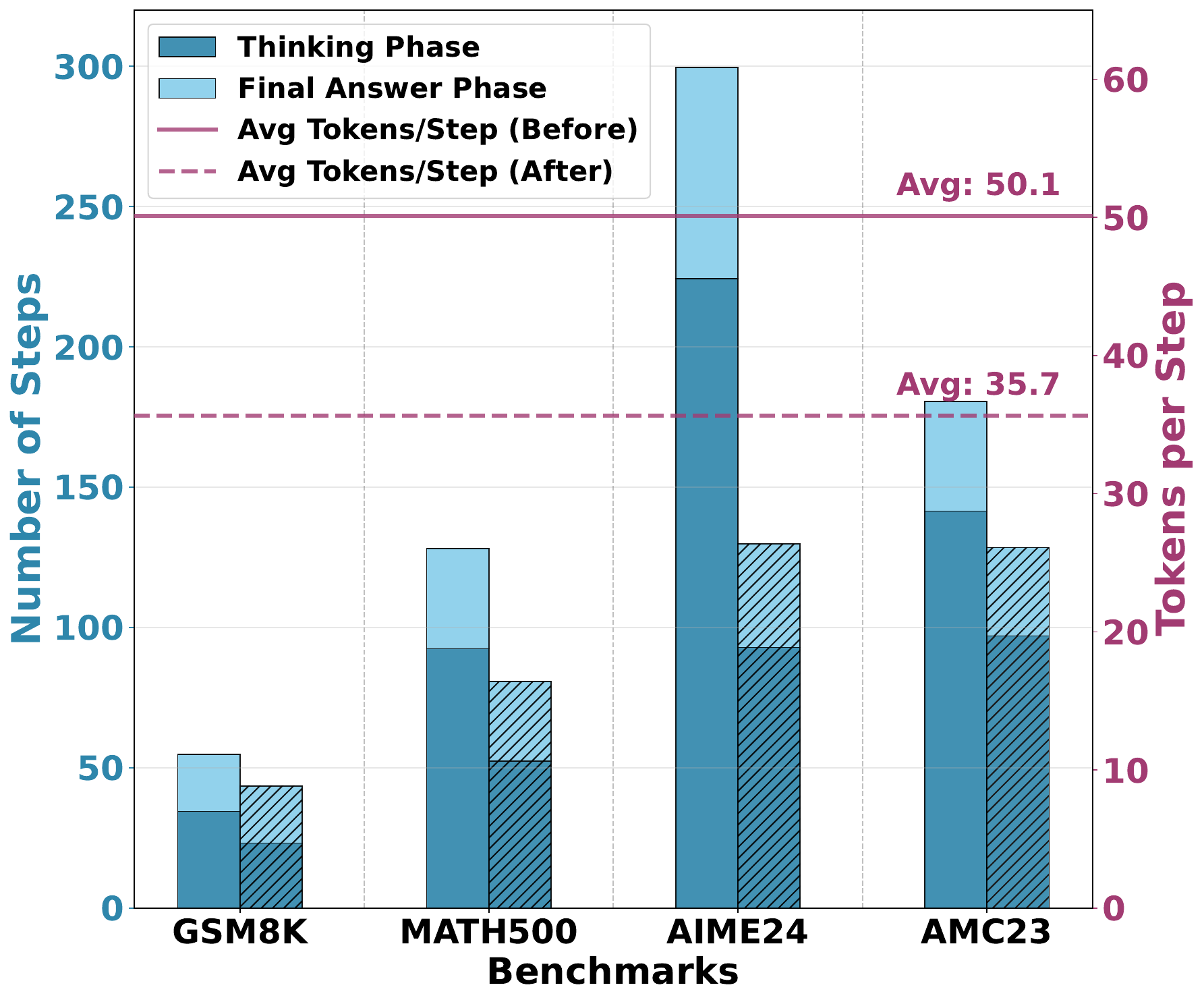}
    \label{fig:qwen3_4b_training_comparison}
  \end{minipage}%
  \vspace{-5mm}
  \caption{(a) Entropy changes before and after training with PEAR across thinking and final answer phases. (b) Changes in the number of reasoning steps and average tokens per step for Qwen3-4B. PEAR reduces both the number of reasoning steps and the average tokens per step.}
  \label{fig:result_entropy_and_step_tokens}
  \vspace{-5mm}
\end{figure}

We further analyze how PEAR influences model reasoning across different phases, focusing on changes in entropy, number of reasoning steps, and average tokens per step after training with PEAR.

As shown in Figure~\ref{fig:result_entropy_and_step_tokens}(a), PEAR consistently reduces the overall entropy across all evaluated models. 
Crucially, the largest reduction occurs in the thinking phase, where excessive exploration had previously contributed to unnecessarily long reasoning traces. 
This demonstrates that our reward effectively steers models toward more confident and focused reasoning, eliminating redundant exploratory steps in the thinking process. 
In contrast, the final answer phase shows a slight increase in entropy, indicating that the model retains flexibility when articulating its conclusions. 
Such phase-specific adjustments highlight PEAR’s ability to suppress over-exploration during reasoning while still supporting diversity and completeness in the final answer through the control towards entropy.

Figure~\ref{fig:result_entropy_and_step_tokens}(b) 
illustrates the changes in the number of reasoning steps and tokens per step for the Qwen3-4B model across all benchmarks before and after applying PEAR. 
The results show that PEAR not only reduces the total number of reasoning steps but also decreases the average tokens per step, reflecting a shift toward more deterministic and efficient reasoning. 
Importantly, the reduction is concentrated in the thinking phase, consistent with PEAR’s objective of discouraging excessive exploration while maintaining entropy in the final answer phase. 
This effect is especially pronounced on more challenging datasets such as AIME24, where the number of thinking steps is reduced by more than half. 
These results further validate the effectiveness of PEAR in producing concise reasoning trajectories without compromising solution quality.

Crucially, these findings explain why PEAR achieves substantial reductions in response length without sacrificing accuracy, highlighting phase-dependent entropy as a powerful control signal for balancing efficiency and performance in large reasoning models.

\subsection{Hyperparameter study}

\begin{wrapfigure}{r}{0.45\linewidth} 
  \vspace{-5mm}
  \centering
  \includegraphics[width=0.95\linewidth]{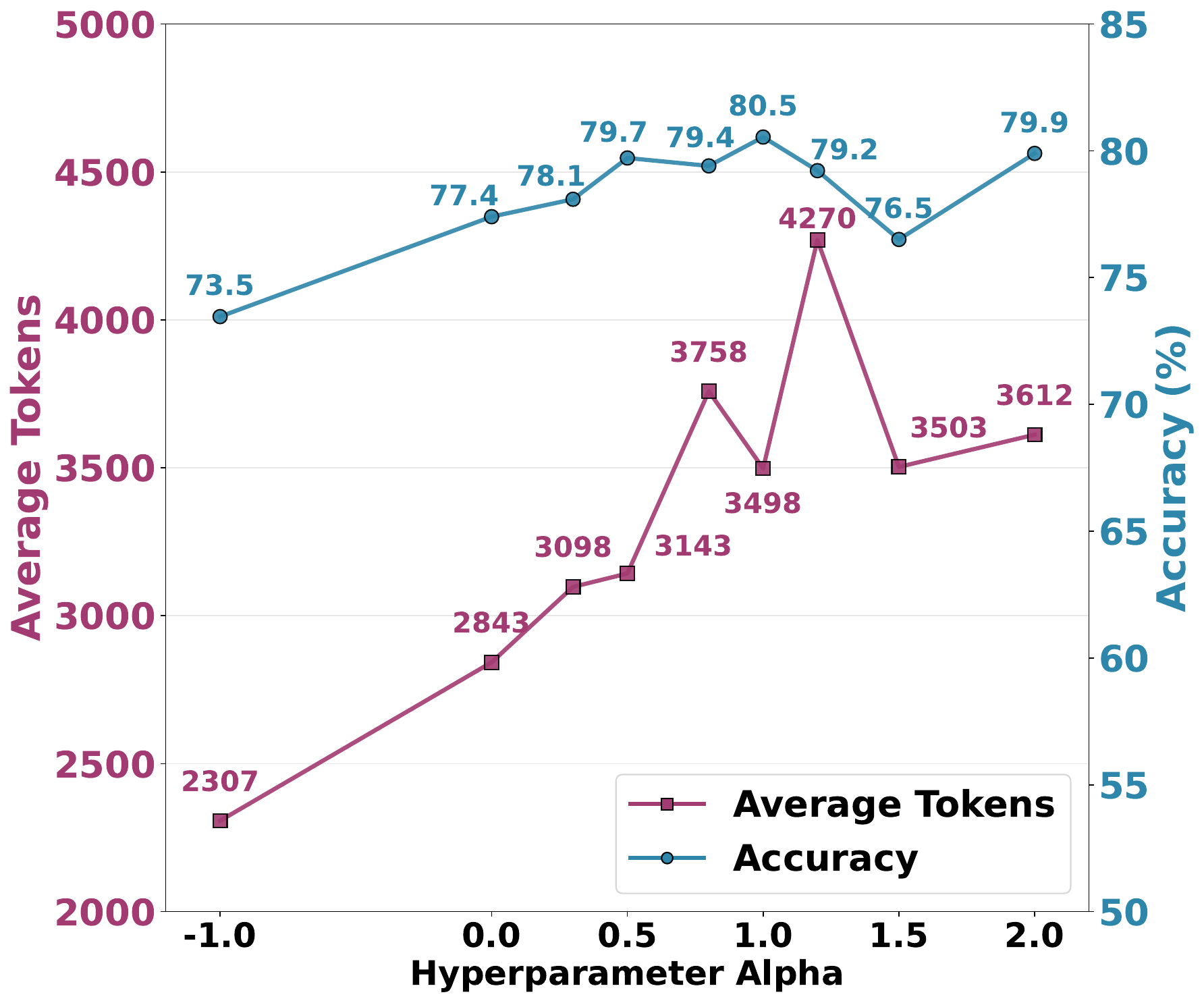}
  \vspace{-2mm}
  \caption{Average accuracy and response length of Qwen3-4B trained with different $\alpha$.}
  \label{fig:hyperparameter_Qwen3-4B}
  \vspace{-4mm}
\end{wrapfigure}

A central hyperparameter in our reward design is the coefficient $\alpha$ for final answer phase's entropy.
This parameter directly controls the extent to which the model is encouraged for higher entropy in the final answer phase.
Figure~\ref{fig:hyperparameter_Qwen3-4B} illustrates the impact of the hyperparameter $\alpha$ on Qwen3-4B across four benchmarks. 
By default, $\alpha$ is set to a positive value in order to avoid ``reward gaming'', where the model drives entropy down indiscriminately to maximize reward, which often leads to degraded performance. 

The experiments confirm this hypothesis. 
When \mbox{$\alpha=0$}, post-thinking entropy is ignored, and the model is optimized solely to minimize entropy in the thinking phase. 
While efficient, this strict reduction harms accuracy, as the model loses the flexibility needed in the answer phase to refine or adjust its predictions. 
The problem becomes even more pronounced when \mbox{$\alpha=-1$}, where both the reasoning and answer phases are simultaneously penalized for entropy. 
In this setting, the model is overly constrained, producing shorter but less reliable responses and further degrading performance. 

As $\alpha$ increases, the penalty on post-thinking entropy becomes stronger. 
This relaxes the restrictive effect on the answer phase, allowing the model to preserve higher entropy where needed and thereby improving accuracy. 
At moderate values of $\alpha$ (e.g., $1$), we observe a favorable balance: the model reduces redundancy in its reasoning while maintaining strong performance. 
However, when $\alpha$ is set too high, the penalty effect becomes negligible, and the model’s behavior converges toward the baseline, producing longer responses and diminishing the efficiency gains.

%% file: Tables/main_results_clear.tex
\begin{table}[t]
\centering
\scriptsize
\setlength{\tabcolsep}{3pt}
\renewcommand{\arraystretch}{1.15}
\caption{Acc@1 results on four mathematical reasoning benchmarks across three LRMs. 
$\downarrow$ indicates the relative change with respect to the \emph{Original} row of each model. 
PEAR consistently achieves the largest reduction in token usage across model scales, while maintaining comparable accuracy.}
\resizebox{0.95\linewidth}{!}{%
\begin{tabular}{l|cc|cc|cc|cc|cc}
\toprule
\multirow{2}{*}{Method}
& \multicolumn{2}{c|}{\textbf{GSM8K}}
& \multicolumn{2}{c|}{\textbf{MATH500}}
& \multicolumn{2}{c|}{\textbf{AIME24}}
& \multicolumn{2}{c|}{\textbf{AMC23}}
& \multicolumn{2}{c}{\textbf{Average}} \\
\cmidrule(lr){2-3}\cmidrule(lr){4-5}\cmidrule(lr){6-7}\cmidrule(lr){8-9}\cmidrule(lr){10-11}
& Acc & Tok & Acc & Tok & Acc & Tok & Acc & Tok & Acc & Tok \\
\midrule
\rowcolor{gray!15}
\multicolumn{11}{l}{\textbf{\textit{\small DeepSeek-R1-Distill-Qwen-1.5B}}} \\
Original & 85.97 & 1496 & 75.00 & 3620 & 26.66 & 8843 & 70.00 & 5253 & 64.41 & 4853 \\
GRPO & 87.86 & 1493 & 76.80 & 3132 & 33.33 & 7839 & 67.50 & 4899 & 66.37 & 4341~{\color{blue}\scriptsize($\downarrow$ 10.6\%)} \\
Step Entropy & 85.59 & 1629 & 76.80 & 3298 & 26.66 & 5640 & 70.00 & 4911 & 64.76 & 3870~{\color{blue}\scriptsize($\downarrow$ 20.3\%)} \\
LCPO & 87.11 & 2149 & 76.00 & 2895 & 26.66 & 5358 & 70.00 & 3324 & 64.94 & 3432~{\color{blue}\scriptsize($\downarrow$ 29.3\%)} \\
PEAR & 87.94 & 624 & 77.20 & 2358 & 23.33 & 5379 & 70.00 & 3705 & 64.62 & 3016~{\color{blue}\scriptsize($\downarrow$ 37.8\%)} \\
\midrule
\rowcolor{gray!15}
\multicolumn{11}{l}{\textbf{\textit{\small Qwen3-4B}}} \\
Original & 94.69 & 2634 & 85.40 & 5795 & 56.66 & 16792 & 87.50 & 9234 & 81.06 & 8614 \\
GRPO & 94.38 & 2321 & 84.80 & 5434 & 63.33 & 14061 & 90.00 & 8568 & 83.13 & 7596~{\color{blue}\scriptsize($\downarrow$ 11.8\%)} \\
Step Entropy & 94.84 & 2261 & 85.40 & 4704 & 60.00 & 9467 & 87.50 & 7317 & 81.93 & 5937~{\color{blue}\scriptsize($\downarrow$ 31.1\%)} \\
LCPO & 93.47 & 1846 & 84.20 & 3569 & 63.33 & 8528 & 85.00 & 6518 & 81.50 & 5115~{\color{blue}\scriptsize($\downarrow$ 40.6\%)} \\
PEAR & 94.01 & 1439 & 84.00 & 2695 & 56.66 & 5685 & 87.50 & 4173 & 80.54 & 3498~{\color{blue}\scriptsize($\downarrow$ 59.4\%)} \\
\midrule
\rowcolor{gray!15}
\multicolumn{11}{l}{\textbf{\textit{\small Qwen3-8B}}} \\
Original & 96.13 & 2335 & 86.60 & 5532 & 63.33 & 14977 & 90.00 & 8161 & 84.02 & 7751 \\
GRPO & 95.83 & 1999 & 85.20 & 5375 & 66.66 & 13195 & 90.00 & 7881 & 84.42 & 7113~{\color{blue}\scriptsize($\downarrow$ \textcolor{white}{0}8.2\%)} \\
Step Entropy & 95.14 & 2087 & 86.00 & 4658 & 60.00 & 6816 & 90.00 & 7352 & 82.79 & 5228~{\color{blue}\scriptsize($\downarrow$ 32.6\%)} \\
LCPO & 94.54 & 1645 & 85.00 & 4234 & 63.33 & 7173 & 82.50 & 6961 & 81.34 & 5003~{\color{blue}\scriptsize($\downarrow$ 35.5\%)} \\
PEAR & 94.54 & 1092 & 85.40 & 2664 & 60.00 & 6104 & 92.50 & 4045 & 83.11 & 3476~{\color{blue}\scriptsize($\downarrow$ 55.2\%)} \\
\bottomrule
\end{tabular}%
}
\label{tab:main_results_clear}
\vspace{-4mm}
\end{table}

%% file: Sec/02_Related_Work_Concise.tex
\section{Related Work}
\subsection{Efficient Reasoning}
A growing body of research has focused on improving the efficiency of LRMs. 
Early exit stops model dynamically once certain criteria has been reached \citep{liao2025fractured}.
Typical methods include designing stopping rules based on internal reasoning state \citep{yang2025dynamic, qiao2025concise, zhu2025think, xu2025scalable}, generation behavior \citep{wang2025wait, wang2025thoughts, liu2025answer}, or without relying on pre-defined triggers \citep{dai2025s}. 
Another complementary research direction focuses on compressing chain-of-thought reasoning traces, such as parallel thinking compression \citep{munkhbat2025self, ghosal2025does}, filtering or summarizing intermediate reasoning tokens and steps \citep{yu2025long, luo2025autol2s, yuan2025not, xia2025tokenskip, zhao2025let}, and compression reward mechanisms \citep{cheng2025optimizing, zeng2025done}. Notably, \citet{li2025compressing} introduce step entropy for quantifying the informational contribution of each reasoning step within CoT trajectories, enabling selective removal of low-entropy steps.
Besides, adaptive reasoning methods attempt to dynamically adjust the depth or length of reasoning depending on the difficulty of the input, this includes carefully designed reward \citep{jiang2025think, wang2025adaptive, luo2025adar1} and reasoning mode switching \citep{zhang2025othink, huang2025adactrl, zhang2025adaptthink}. 
For example, LCPO \citep{aggarwal2025l1} include user-specified length constraint into the training reward to guide the model toward answering within the constraint. 
However, such methods discard valuable intermediate reasoning that could improve accuracy. In contrast, our method utilizes the intrinsic phase-dependent entropy as reward signal, making it an adaptive and model-driven approach to helps the model reason more efficiently.

\subsection{Reasoning Through Entropy Control}
With the increasing research focus on Reinforcement Learning with Verifiable Rewards (RLVR), model entropy \citep{shannon1948mathematical} has emerged as a powerful internal signal for shaping reasoning behaviors in large language models. 
Recent work has investigated how policy entropy evolves during reinforcement learning-based post-training of reasoning models. \citet{zhang2025no} reveal the correlation between entropy collapse and performance saturation as well as subsequent degradation. 
\citet{cui2025entropy} further shows how high-probability/high-advantage updates systematically reduce entropy. 
Another complementary direction treats entropy minimization itself as supervision by directly minimizing token-level entropy via finetuning or using negative entropy as the sole reward in RL \citep{agarwal2025unreasonable, prabhudesai2025maximizing}. 
Besides, recent work has explored augmenting reinforcement learning approaches by incorporating entropy-based mechanisms to encourage exploration in reasoning chains \citep{zhang2025edge, cheng2025reasoning}. 
Furthermore, \citet{wang2025beyond} reveal that the effectiveness of RLVR stems primarily from optimizing high-entropy tokens that determine critical reasoning directions. Selectively targeting these high-entropy minority tokens during optimization can substantially enhance reasoning capabilities while improving computational efficiency.
While most existing studies leverage entropy to improve reasoning capability, our approach uses entropy as a control signal for efficiency, enabling adaptive length control without explicit token budgets while preserving accuracy. This reframes entropy not only as a tool for capability shaping but also as a principled knob for controlling the cost of reasoning.

%% file: Sec/07_Conclusion.tex
\section{Conclusion}
In this work, we conduct empirical analysis and observed the consistent positive relation between entropy and response length across reasoning stages: the thinking phase exhibits higher entropy, reflecting exploratory behavior of longer response, while the final answer phase shows lower entropy, indicating more deterministic solution. 
Based on this finding, we address the challenge of efficient reasoning by introducing Phase Entropy Aware Reward (PEAR), a reward mechanism that distinguishes entropy between thinking phase and final answer phase during training. By discouraging entropy in thinking phase while preserving flexibility in final answer phase, PEAR enables adaptive control of response length without requiring explicit length targets or rigid truncation rules.
Extensive experiments across four benchmarks have demonstrated that PEAR reduces token redundancy by a large percentage of 37.8\% to 59.4\% while preserving accuracy. Besides, PEAR also demonstrate strong generalization capability to out-of-distribution tasks.

%% file: Appendix/02_Qwen3_8B_Entropy_filter_results.tex
\section{Entropy Filtering Experiments for Qwen3-8B}
\label{app:entrop_filter_Qwen3_8B}
\begin{figure}[t] 
  \centering
  \includegraphics[width=0.9\columnwidth]{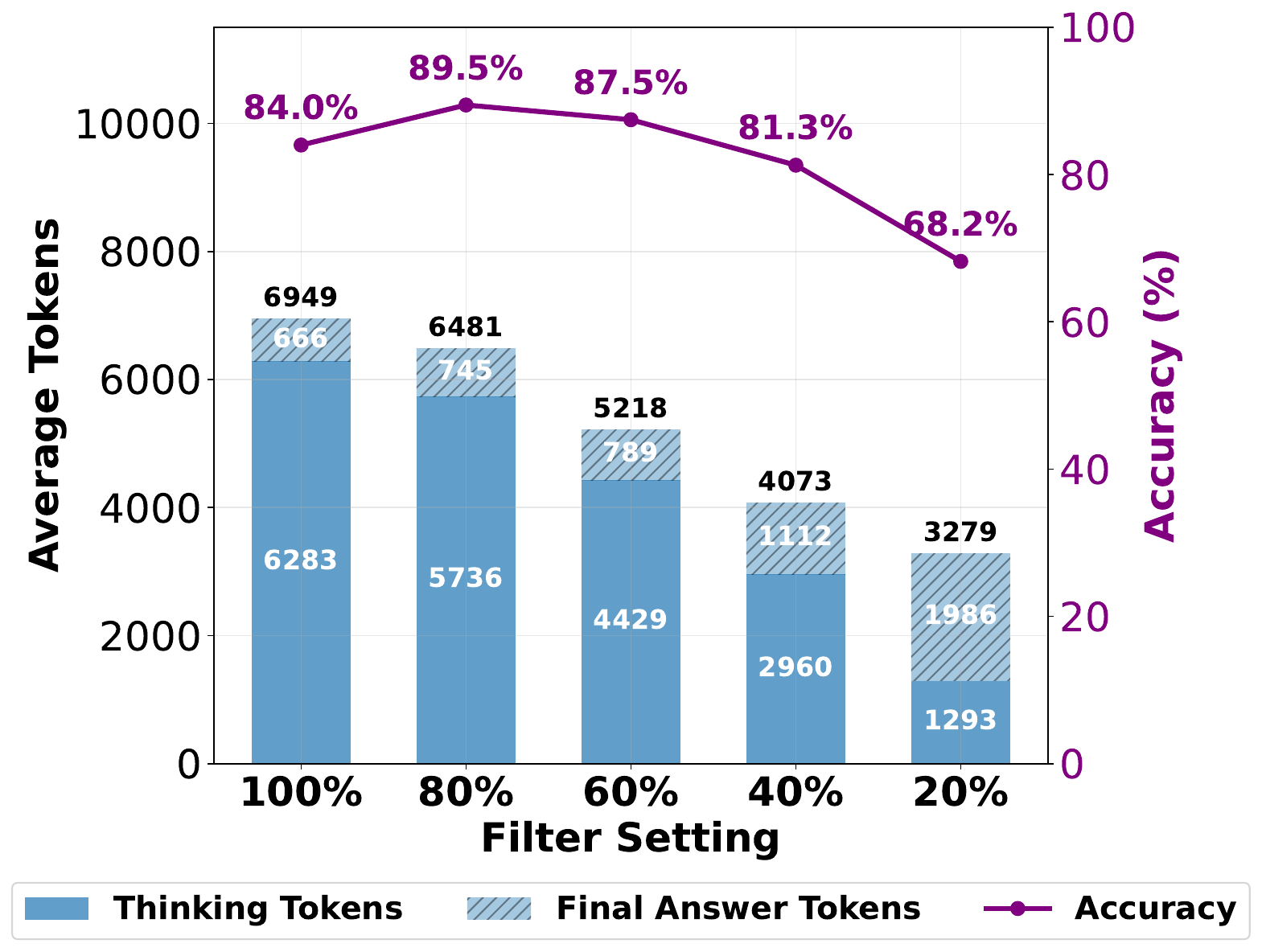} 
  \caption{Accuracy and average response length in the entropy filtering experiments on Qwen3-8B.}
  \label{fig:entropy_filter_Qwen3_8B}
\end{figure}

Figure~\ref{fig:entropy_filter_Qwen3_8B} demonstrates the entropy filtering experiment result on Qwen3-8B.
The results reveal a similar trend as Qwen3-4B discussed in Section~\ref{sec:entropy_filter}. 
When retaining 80\% or 60\% of low-entropy tokens, accuracy remains stable or even improves compared to the unfiltered baseline. 
Performance degradation only emerges under more aggressive filtering: retaining 40\% or fewer low-entropy tokens leads to a sharp drop in accuracy, showing that essential reasoning steps are lost when the trace is compressed too heavily. Notably, the length of the final answer phase also remains relatively unchanged across filtering levels, reinforcing that redundancy is concentrated in the thinking phase.

This result further supports the conclusion that the high-entropy tokens being removed mainly drive excessive exploration rather than contributing to correct reasoning, and their absence reduces noise in the reasoning process.

%% file: Appendix/03_Experiment_Details_for_baselines.tex
\section{Experiment details for baseline methods}
\label{app:baseline method}

We evaluate three baseline methods: \textbf{GRPO} (Group Relative Policy Optimization) \citep{shao2024deepseekmath}, 
\textbf{Step Entropy} \citep{li2025compressing}, and \textbf{LCPO} (Length-Controlled Policy Optimization) \citep{aggarwal2025l1} using the GSM8K training set \citep{cobbe2021training}. 
Experiments are conducted across three model sizes: DeepSeek-R1-Distill-Qwen-1.5B \citep{guo2025deepseek}, Qwen3-4B, and Qwen3-8B \citep{yang2025qwen3}. 
The implementation details for each baseline are provided below.

For GRPO \citep{shao2024deepseekmath}, we use the open-source \texttt{verl} framework \citep{sheng2025hybridflow}\footnote{\url{https://github.com/volcengine/verl}} with the original rule-based reward, which assigns a reward of 1 for correct answers and 0 otherwise. 
We set the rollout number to 8 and the KL penalty coefficient to $1\times10^{-3}$.

For Step Entropy \citep{li2025compressing}, we use the official implementation provided by the authors\footnote{\url{https://github.com/staymylove/COT_Compression_via_Step_entropy}}. 
The method follows a two-stage training strategy: Supervised Fine-Tuning (SFT) with pruned CoT data, followed by Reinforcement Learning (RL) with GRPO. 
During the SFT stage, training is performed with mixed precision (FP16), a learning rate of $2\times10^{-5}$, and a weight decay of 0.01. 
In the RL stage, the learning rate is set to $1\times10^{-5}$ and the KL penalty is fixed at 0.1.

For LCPO \citep{aggarwal2025l1}, we use the official codebase provided by the authors\footnote{\url{https://github.com/cmu-l3/l1}} and follow the L1-Exact setup. 
Training is performed with GRPO under length control and a maximum length constraint. 
We set the learning rate to $1\times10^{-6}$ with a batch size of 64, and restrict the context length to 4K tokens during training. 
Rollout number is fixed at 8 with a sampling temperature of 0.6, and the KL penalty coefficient is set to $1\times10^{-3}$.

%% file: Appendix/04_Evaluation_Benchmarks.tex
\section{Evaluation Benchmarks}
\label{app:benchmarks}
To evaluate the effectiveness and generalizability of our compression method, we benchmark on four standard mathematical reasoning datasets.

\textbf{GSM8K test set} \citep{cobbe2021training} is a carefully designed benchmark comprising 1,319 grade-school mathematics word problems. 
Each question typically requires two to eight sequential reasoning steps, primarily involving basic arithmetic operations applied across multiple intermediate stages. 
\textbf{MATH500} \citep{hendrycks2021measuring} contains a subset of 500 problems drawn from high school mathematics competitions. 
We follow the evaluation setup of OpenAI by adopting the same curated subset. 
\textbf{AIME24} \citep{li2024numinamath} features 30 problems from the 2024 American Invitational Mathematics Examination (AIME). 
As one of the most prestigious secondary-level competitions, AIME problems demand sophisticated reasoning across diverse topics, including algebra, combinatorics, geometry, number theory, and probability. 
\textbf{AMC23} \citep{li2024numinamath} consists of 40 problems taken from the 2023 American Mathematics Competition (AMC). The dataset covers core high school mathematics domains such as algebra, geometry, combinatorics, and number theory, providing a broad yet rigorous evaluation of mathematical reasoning ability.